\def\year{2020}\relax
\documentclass[letterpaper]{article} 
\usepackage{aaai20}  
\usepackage{times}  
\usepackage{helvet} 
\usepackage{courier}  
\usepackage[hyphens]{url}  
\usepackage{graphicx} 
\usepackage{graphicx}  
\title{From Natural Language Instructions to Complex Processes: \\
Issues in Chaining Trigger Action Rules}
\author{Nobuhiro Ito\textsuperscript{\rm 1} {\rm and}
Yuya Suzuki\textsuperscript{\rm 1} {\rm and}
Akiko Aizawa\textsuperscript{\rm 2,}\textsuperscript{\rm 1}\\
\textsuperscript{\rm 1}The University of Tokyo \hspace{5pt}
\textsuperscript{\rm 2}National Institute of Informatics \\
nobuhiro.ito@is.s.u-tokyo.ac.jp \hspace{5pt} \{ysuzuki, aizawa\}@nii.ac.jp}

\begin{document}

\maketitle

\begin{abstract}
Automation services for complex business processes usually require a high level of information technology literacy. There is a strong demand for a smartly assisted process automation (IPA: intelligent process automation) service that enables even general users to easily use advanced automation. A natural language interface for such automation is expected as an elemental technology for the IPA realization. The workflow targeted by IPA is generally composed of a combination of multiple tasks. However, semantic parsing, one of the natural language processing methods, for such complex workflows has not yet been fully studied. The reasons are that (1) the formal expression and grammar of the workflow required for semantic analysis have not been sufficiently examined and (2) the dataset of the workflow formal expression with its corresponding natural language description required for learning workflow semantics did not exist. This paper defines a new grammar for complex workflows with chaining machine-executable meaning representations for semantic parsing. The representations are at a high abstraction level. Additionally, an approach to creating datasets is proposed based on this grammar.
\end{abstract}

\section{Introduction}
Automation services for business process like robotic process automation (RPA) have recently attracted increasing attention. These automation services typically require advanced information technology literacy when creating automation programs. Thus, it is difficult for non-skilled users to make use of the services. A smart interface that can create and execute an automated program specified in a natural language (NL) description would be useful. This smart interface is one of the intelligent process automation (IPA) realizations. We focus on a natural language processing method that plays an important role in this interface: semantic parsing. Semantic parsing consists of three components: (a) a natural language description, (b) a machine-executable meaning representation (MEMR), and (c) a parser that converts (a) to (b). MEMR can be a formal expression that follows a specific grammar. Workflows targeted by automation systems represented by RPA are generally complex workflows composed of a combination of multiple tasks. In studies on semantic parsing, (1) the formal expression and the formal grammar for expressing such workflows have both not been sufficiently examined. As a result, (2) there was no dataset that had a pairing of a complex workflow with its corresponding NL description. 

\subsection{Formal Expression and Grammar}
With respect to the expression and the grammar, we focus on two problems: an expression unit, and a formal grammar for complex workflows. 
\subsubsection{Expression Unit}
Several types of expressions exist depending on the abstraction level of the expression. Many studies have been conducted on parsing NL description to code, such as transition-based neural semantic parsing like TranX \cite{Yin2017,Yin2018b}. The field of code generation typically uses expressions with low abstraction: code itself. Therefore, the expression becomes bloated when attempting to express a complex workflow that is composed of hundreds of lines of code. Consider an expression with high abstraction that is much closer to human language, as exemplified by calling macros, APIs, and modules, \footnote{This expression can be considered as a ``no-code/low-code'' expression that has recently been refocused in RPA.} such as ``Insert Rows Into a Google Spreadsheet''; this is a straightforward approach to avoiding the bloating problem. An expression with high abstraction is typically used in a trigger action program (TAP) format that performs a certain action when a certain trigger occurs. A notable TAP dataset is the IFTTT dataset \cite{Ur2016,Mi2017}. However, the workflow targeted by the IFTTT dataset itself is too simple to use for complex workflows. 
\subsubsection{Formal Grammar}
\begin{figure}[htb]
\centering
\includegraphics[width=8.5cm]{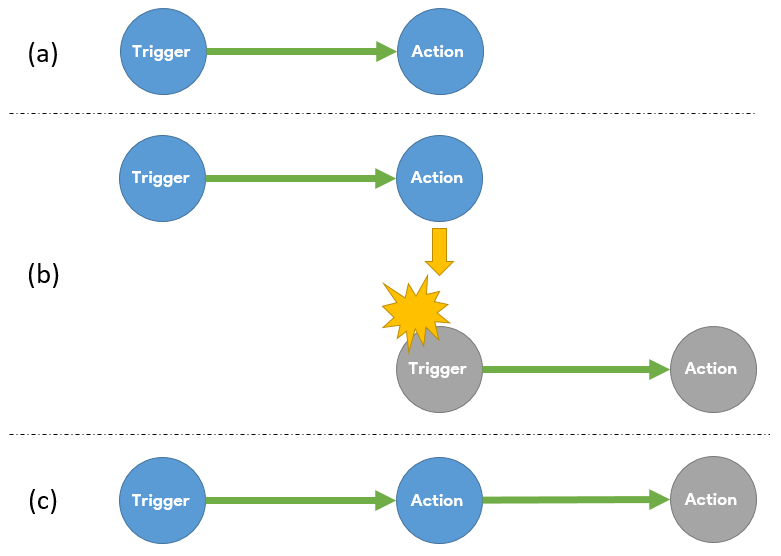}
\caption{TAP Chain}
\label{single_tap_vs_tap_chain}
\end{figure}

Although it is necessary to express a series of processes as a whole in one shot, collecting such datasets is inherently difficult. We focus on the form of the TAP chain in formulating a formal grammar. TAP can be considered a type of MEMR. The advantage of TAP is that it expresses the essence of being event-driven in a straightforward manner, so it has high versatility. Moreover, there are datasets and existing studies on TAP. Complex workflows are composed of process chains like ``If this triggered, then do this action and do this action separately, and finally do this action.'' On the other hand, a single TAP only ends with ``If this triggered, then do this action,'' ((a) in Figure \ref{single_tap_vs_tap_chain}) but we can assume that performing this action causes another trigger ((b) in Figure \ref{single_tap_vs_tap_chain}). Thus, we can express a complex workflow by repeating TAP chaining ((c) in Figure \ref{single_tap_vs_tap_chain}). We call this a ``TAP chain'' and incorporate it into the grammar formulation\footnote{It seems important to consider the concept of type or category of a MEMR. The type may include semantic content indicating the close connectivity of each task.}. 
This facilitates conversion to complex workflows through relatively simple MEMRs.\footnote{In this paper, we use the TAP chain-based formal expression for workflow for the reason already described, but other formal expressions for workflow can be also considered. We reserve this issue for future work.} Especially, this enables us to apply the already-mentioned transition-based neural semantic parsing to a workflow intricately combined with such MEMRs. Transition-based neural semantic parsing does not generate code directly, but generates a sequence of grammars (in the form of context-free grammars)\footnote{TranX mainly targets a low abstraction level. The main focus of TranX for applying the parsing is to ensure that the parsing results are legal in the context of the Python abstract grammar.}. To the best of our knowledge, no such study has been conducted thus far because of the lack of an adequate grammar for such workflows and parsing.

\subsection{Complex Workflow Dataset}
In our creation of complex workflow dataset, we use the IFTTT dataset that contains TAP programs with their corresponding NL descriptions. From the IFTTT dataset, we manually extract the relationships where one TAP triggers another TAP, and generate the TAP chain rules. After that, a TAP chain rule is randomly applied to generate a complex workflow. Since it is applied randomly, the usefulness of the generated workflow is manually annotated. 
NL description corresponding to the created workflow is generated by fusing NL description of each TAP that constitutes the workflow. Two approaches are considered: rule-based generation, and sentence fusion generation (one approach in abstractive summarizations). The generated NL descriptions are manually annotated especially with respect to test data.

\vspace{\baselineskip}
The main contribution of this study is the definition of a new grammar for semantic parsing to complex workflows. In addition, an approach to creating the dataset is proposed based on this grammar. In the following, we concisely explain the related work, then propose workflow patterns grammar (WPG) and dataset creation, then concisely refer to the model, and finally conclude this short paper.

\section{Related Work}
\subsection{Semantic Parsing}
The process of converting NL into MEMR is known as semantic parsing. A typical example of semantic parsing is SQL generation for database queries. A semantic parser translates the sentence ``How many people live in Seattle?'' to ``{\sf SELECT Population FROM CityData where City==Seattle}". Then the SQL query is executed to obtain the correct answer, ``620,778'' \cite{gardner-etal-2018-neural}. Another typical example of semantic parsing is code generation where a single function declaration or class declaration is viewed as an MEMR \cite{gardner-etal-2018-neural}. Rather than generating code directly, transition-based neural semantic parsing like TranX generates a sequence of ASDL grammars that are sequentially expanded and applied to generate MEMRs \cite{Yin2017,Yin2018b}. However, these parsing models mainly target a low abstraction level of formal expression.

\subsection{TAP/IFTTT}
A typical web service of TAP is the IFTTT \cite{Ur2016,Mi2017}. As an example, the NL description ``youtube upload to blogger new post'' is converted into an MEMR with a high abstraction level TRIGGER ``{\sf YouTube.New\_public\_video\_uploaded\_by\_you}'' and ACTION ``{\sf Blogger.Create\_a\_post}.'' A study of simple grammar-based semantic parsing without neural model has been conducted \cite{Quirk2015}. Additionally, a method using neural semantic parsing has been proposed \cite{Beltagy2016,Chen2016,Dong2016,Dong2019}. Furthermore, a dialogue model \cite{Chaurasia2017} and a reinforcement learning dialogue model \cite{Yao2019} have also been proposed. However, a target workflow of IFTTT itself is too simple to be able to directly automate complex real-world workflows.

\section{Workflow Patterns Grammar}
In this paper, we define a new grammar for semantic parsing to complex workflows (Table \ref{WPG}). This enables us to apply the transition-based neural semantic parsing to a workflow intricately combined with MEMRs. The specific notation follows \citeauthor{Yin2018b} \shortcite{Yin2018b}, which mainly refer to the Python ASDL grammar: The notation ``?'' represents the optional type, which can have one value or a null value, and the notation ``*'' indicates the sequential type, which can have two or more values.
\begin{table}[htb]
\centering
\scalebox{0.75}
{
\begin{tabular}{cclc}
\hline
\\
{\it stmt} &$=$& Workflow({\it wpg} pattern) &$\cdots$(1)\\
{\it wpg} &$=$& Sequence({\it func}? trigger, {\it func} action) &\\
&$|$& Parallel\_Split({\it func}? trigger, {\it func}* action) &$\cdots$(2)\\
{\it func} &$=$& Call({\it type} channel, {\it wpg}? next) &$\cdots$(3)\\
{\it type} &$=$& Type\_A $|$ Type\_B $|$ Type\_C ... &$\cdots$(4)\\        \\
\hline
\end{tabular}
}
\caption{Workflow Pattern Grammar}
\label{WPG}
\end{table}

\subsection{Workflow Initialization}
The first line (1) in the table indicates the start point of the workflow generation. The {\it stmt} type evokes a constructor with the {\it wpg} type argument called ``pattern.'' This expansion delivers the start point of the workflow.

\subsection{Workflow Patterns}
In workflows like the ones in office workplaces, there are patterns that repeatedly occur. \citeauthor{Russell2016} \shortcite{Russell2016} introduced five basic patterns to capture the elementary aspects of the flow: sequence, parallel split, synchronization, exclusive choice, and simple merge. In this study, we consider two patterns, sequence and parallel split, to maintain the tree structure and simplicity of the MEMRs. These constructors are shown in line (2) of Table \ref{WPG}. The first pattern expands {\it wpg} to a sequence pattern: that is, it evokes the sequence pattern constructor. The second pattern expands {\it wpg} to a parallel split pattern: that is, it evokes the parallel split pattern constructor. 

WPG expression in this paper also considers the flow of processed data. For example, the ``Send Text to Me'' function has no return value and therefore no function to connect to the next. Since the ``Archive Text in Spread Sheet'' function must receive text data, it cannot follow a ``Send Text to Me'' function having no return. If the processed output data from functions are different, it is straightforward to handle them in different branches. Also, transition-based neural semantic parsing learns the sequence of grammar expansions corresponding to NL descriptions. In this learning, the parallel split is expected to work as a signal token to decide whether subsequent flow should branch, based on its previous trigger or action and NL descriptions.

\subsection{TAP Chaining}
As already mentioned, a complex workflow is generated from a simple MEMR by considering a TAP chain, as presented in line (2) in Table \ref{WPG}. The sequence pattern's arguments are the {\it func}? type argument called ``trigger'' and the {\it func} type argument called ``action.'' Because this pattern is simple, such that when the ``trigger'' is evoked, the ``action'' is activated, we consider this pattern to have two arguments. The notation ``?'' represents the optional type, which can have one value or a null value. When two sequence patterns are connected (for example, Sequence(Function A, Function B) and Sequence(Function B, Function C) are connected sequentially), the action in the first sequence pattern is the same as the trigger in the second sequence pattern. Thus, the function is duplicated. We expand the second sequence pattern as Sequence(null, Function C).

The parallel split pattern's arguments are the {\it func}? type argument called ``trigger'' and the {\it func}* type argument called ``action.'' This pattern splits the preceding function's result into two or more functions.

\subsection{TAP Call}
{\it func} is the type that evokes the Call constructor, as presented in line (3) in Table \ref{WPG}. The constructor controls the next task and the next workflow to be executed after the task is completed. The concrete task is derived through expansion from the {\it type} argument called ``channel''. On the other hand, if the task is followed by another task, the constructor should get a value at the {\it type} argument called ``next''. {\it type} is expanded to a concrete macro method class that has concrete functions belonging to the class. This expansion is presented in line (4) in Table \ref{WPG}.

\subsection{Workflow Represented in Abstract Syntax Tree (AST)}
\begin{figure}[htb]
\centering
\includegraphics[width=8cm]{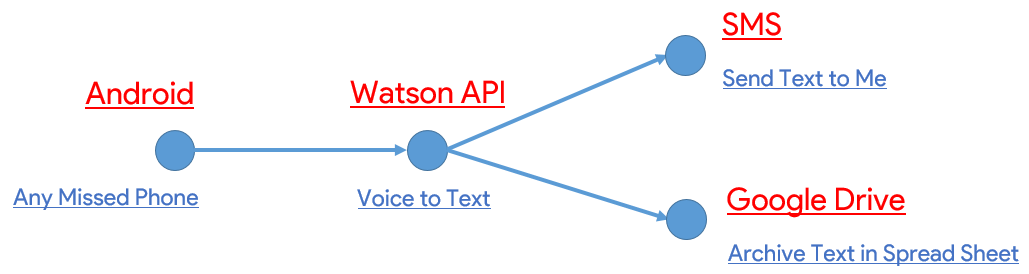}
\caption{Workflow Example}
\label{workflow_example}
\end{figure}

\begin{figure}[htb]
\centering
\includegraphics[width=8cm]{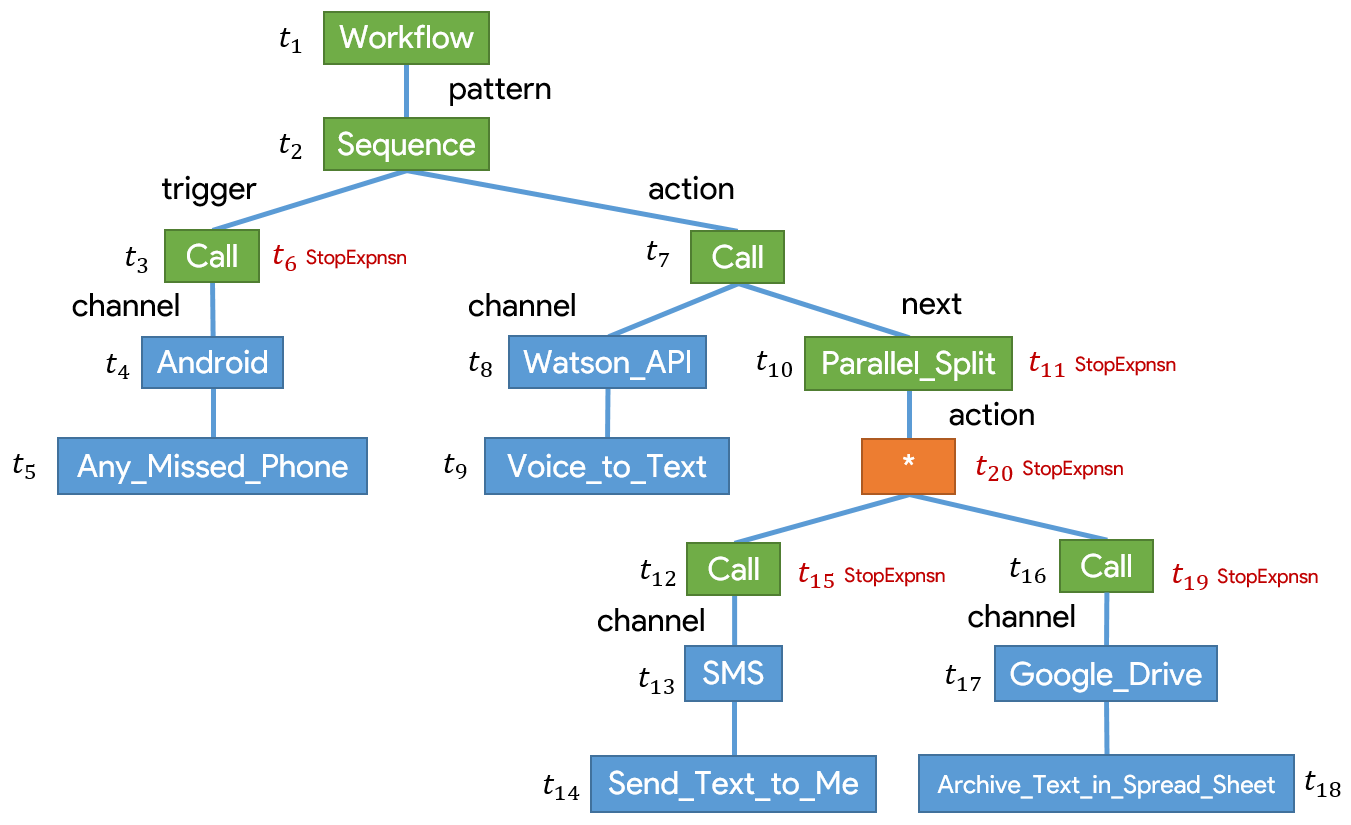}
\caption{Complex Workflow in WAST Form}
\label{w_ast_c}
\end{figure}

Consider a specific example of a workflow represented in AST (WAST). For example, consider a workflow, as depicted in Figure \ref{workflow_example}\footnote{In this paper, to make the discussion easier we use a workflow that is more complex than TAP but still relatively simple. The proposed framework is applicable to versions of workflow that are more complex than this simple example.}.
When WPG is applied to this workflow, the WAST is that in Figure \ref{w_ast_c}.
The WPG grammars applied sequentially are depicted in Table \ref{wpg_c}.
\begin{table}[htb]
\centering
\scalebox{0.75}
{
\begin{tabular}{ccl}
\hline
$t$     &Frontier Field     & Action \\ \hline \hline
$t_1$   &{\it stmt} root    & Workflow({\it wpg} pattern)\\
$t_2$   &{\it wpg} pattern  & Sequence({\it func}? trigger, {\it func} action) \\
$t_3$   &{\it func}? trigger& Call({\it type} channel, {\it wpg}? next) \\
$t_4$   &{\it type} channel & SelectMacr[Android] \\
$t_5$   &Android            & SelectMacr[Any\_Missed\_Phone] \\
$t_6$   &{\it wpg}? next    & StopExpnsn(close the frontier field) \\
$t_7$   &{\it func} action  & Call({\it type} channel, {\it wpg}? next\_wpg) \\
$t_8$   &{\it type} channel & SelectMacr[Watson\_API] \\
$t_9$   &Android            & SelectMacr[Voice\_to\_Text] \\
$t_{10}$&{\it wpg}? next    & Parallel\_Split({\it func}? trigger, {\it func}* action) \\
$t_{11}$&{\it func}? trigger& StopExpnsn(close the frontier field) \\
$t_{12}$&{\it func}* action & Call({\it type} channel, {\it wpg}? next\_wpg) \\
$t_{13}$&{\it type} channel & SelectMacr[SMS] \\
$t_{14}$&SMS                & SelectMacr[Send\_Text\_to\_Me] \\
$t_{15}$&{\it wpg}? next    & StopExpnsn(close the frontier field) \\
$t_{16}$&{\it func}* action & Call({\it type} channel, {\it wpg}? next\_wpg) \\
$t_{17}$&{\it type} channel & SelectMacr[Google\_Drive] \\
$t_{18}$&Google\_Drive      & SelectMacr[Archive\_Text\_Spread\_Sheet] \\
$t_{19}$&{\it wpg}? next    & StopExpnsn(close the frontier field) \\
$t_{20}$&{\it func}* action & StopExpnsn(close the frontier field) \\
\hline
\end{tabular}
}
\caption{WPG Expansion Example}
\label{wpg_c}
\end{table}
The formal expression for parsing is {\sf Sequence ( Android. Any\_Missed \_Phone, Parallel\_Split ( Watson\_API. Voice\_to\_Text, SMS. Send\_Text\_to\_Me, Google\_Drive. Archive\_Text\_in\_Spread\_Sheet))}.

\section{Dataset Creation}
We propose an approach of creating a training dataset and a test dataset for learning transition-based neural semantic parsing for a complex workflow with TAP chain. We basically suppose that the dataset is to be annotated manually. We use an existing TAP dataset which includes corresponding NL descriptions for TAPs. This dataset is beneficial because TAPs and NL descriptions are actually created and used by real users, and the NL descriptions enable the annotator to reuse them to create NL descriptions for the workflows as a whole. 

\subsection{WAST Generation}
\begin{figure}[htb]
\centering
\includegraphics[width=6.5cm]{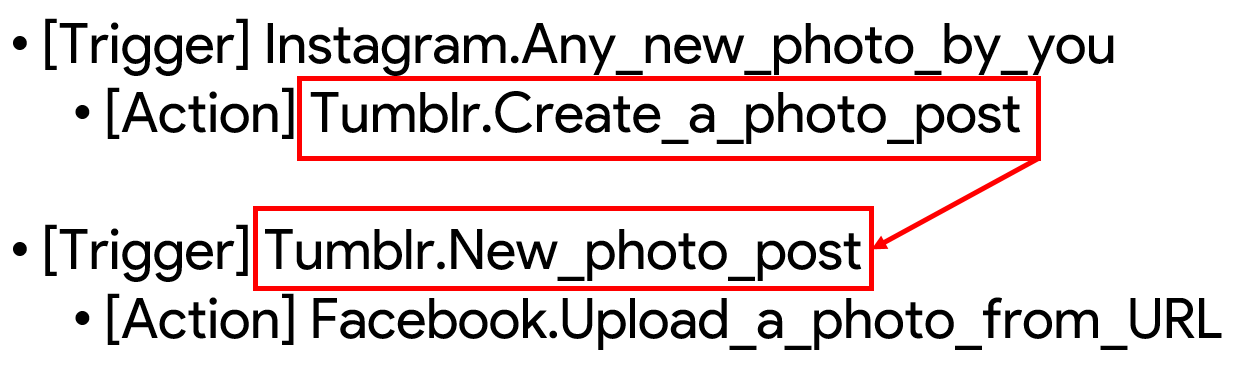}
\caption{TAPs for Vertical Expansion}
\label{vertical_func}
\end{figure}

\begin{figure}[htb]
\centering
\includegraphics[width=6cm]{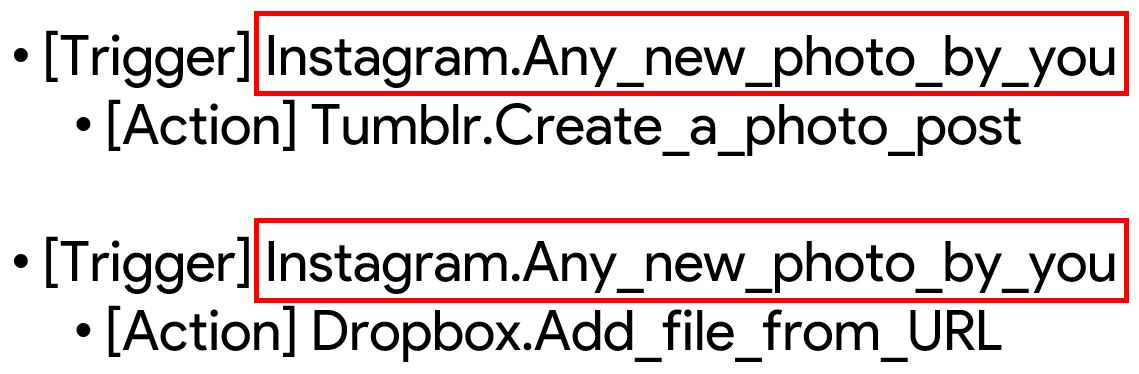}
\caption{TAPs for Horizontal Expansion}
\label{horizontal_func}
\end{figure}

In the IFTTT data, if a trigger function is called, then an action function is invoked. There can exist a case wherein when an action function in one TAP occurs, a trigger function in another TAP is fired simultaneously. We manually conducted such action-evoke-trigger annotations and determined the TAP chaining rules. This chain rule assumes the vertical expansion of TAP in a workflow (Figure \ref{vertical_func}).
On the other hand, the horizontal expansion of TAP is simple; that is, it is sufficient to execute multiple actions that have the same trigger (Figure \ref{horizontal_func}). This makes it possible to create complex workflows from TAP chains.

A complex workflow is generated by randomly chaining TAPs. However, it is unclear whether this automatically generated workflow is really beneficial. Therefore, the automatically generated workflow is annotated in whether: (A) convenient and frequently used, (B) possible to use, or (C) inconvenient and not used. Each TAP that is an element of workflow generation is limited to the TAPs actually created and used by real users. In other words, for each TAP it is assumed that the combination of trigger and action is useful for some user. 

\subsection{Generation of NL instructions}
Furthermore, the graphical form of the automatically generated workflow like Figure \ref{workflow_example} is shown to an annotator, who is asked to annotate the instruction/description that should be given when they ask a machine to perform the workflow\footnote{In the annotation process, we first automatically generate instructions/descriptions by rule-based summarization. Then, annotators review and modify the instruction/description. It is possible to use sentence fusion models \cite{lebanoff-etal-2019-analyzing} to generate these.}. Consequently, a pairing of an NL description and a workflow represented as WAST can be generated.

\section{Model}
With respect to the model, we follow \citeauthor{Yin2018b} \shortcite{Yin2018b}. Transition-based neural semantic parsing has an input of the NL utterance $x$ consisting of $n$ words $\{ w_i \}_{i=1}^{n}$. The parser outputs $a_t$, one of the three transitions: ``ApplyConstr[c] '' expresses the instruction to apply a WPG having constructor $c$, ``SelectMacr'' means to generate a terminal token (function), and ``StopExpnsn'' means stop generating optional or sequential arguments. 

The probability of generating WAST $z$ from NL instructions $x$ is:
\begin{equation}
        q(z|x) = \Pi_t q(a_t|a_{<t}, x).
\end{equation}

The model is trained to maximize the log-likelihood of the transition sequence. Then, the best WAST is inferred from NL description using beam search. 

\section{Discussion}
In this paper, we assume a simple design where each thread progresses independently and focus only on a WAST form with a tree structure. In other words, flows that have branched once never rejoin.  However, complex workflows usually include a simple merge workflow pattern, where branched flows will merge at some point in the following process. WPG needs to be extended to graph a structure. 

This paper referred to general workflow patterns. With the spread of RPA, data on business processes have been accumulated. It is possible that there are common patterns across the companies. Therefore, it is useful to extract such common workflow patterns from real usage data of the RPA products and reduce them to WPG. 

\section{Conclusion}
In this study, we defined a new grammar for chaining high abstraction level MEMRs for semantic parsing into complex workflows. We also proposed an approach to generate a dataset based on this grammar. Consequently, it is expected that an NL interface will be constructed for the complex workflow assumed by IPA. In the future, we intend to perform semantic parsing on the dataset created by this approach.

\bibliographystyle{aaai}
\bibliography{formatting-instructions-latex-2020}

\end{document}